\documentclass{article}

\usepackage[preprint,nonatbib]{neurips_2018}

\usepackage[utf8]{inputenc} 
\usepackage[T1]{fontenc}    
\usepackage{hyperref}       
\usepackage{url}            
\usepackage{booktabs}       
\usepackage{amsfonts}       
\usepackage{nicefrac}       
\usepackage{microtype}      
\usepackage{graphicx} 
\usepackage[ruled,vlined]{algorithm2e} 
\usepackage{amsmath} 
\usepackage{mathdots} 
\usepackage[title]{appendix}
\usepackage[colorinlistoftodos]{todonotes}
\usepackage{todonotes} 
\usepackage{multirow}

\title{Parkinson's Disease Classification via EEG: All You Need is a Single  Convolutional Layer}

\author{%
  Md Fahim~Anjum\thanks{Alternate email: dr.fahim.anjum@gmail.com} \\
  Department of Neurology\\
  University of California San Francisco\\
  San Francisco, CA 94143 \\
  \texttt{fahim.anjum@ucsf.edu} \\
}

\begin{document}

\maketitle

\begin{abstract}
In this work, we introduce LightCNN, a minimalist Convolutional Neural Network (CNN) architecture designed for Parkinson's disease (PD) classification using EEG data. LightCNN's strength lies in its simplicity, utilizing just a single convolutional layer. Embracing Leonardo da Vinci's principle that "simplicity is the ultimate sophistication," LightCNN demonstrates that complexity is not required to achieve outstanding results. We benchmarked LightCNN against several state-of-the-art deep learning models known for their effectiveness in EEG-based PD classification. Remarkably, LightCNN outperformed all these complex architectures, with a 2.3\% improvement in recall, a 4.6\% increase in precision, a 0.1\% edge in AUC, a 4\% boost in F1-score, and a 3.3\% higher accuracy compared to the closest competitor. Furthermore, LightCNN identifies known pathological brain rhythms associated with PD and effectively captures clinically relevant neurophysiological changes in EEG. Its simplicity and interpretability make it ideal for deployment in resource-constrained environments, such as mobile or embedded systems for EEG analysis. In conclusion, LightCNN represents a significant step forward in efficient EEG-based PD classification, demonstrating that a well-designed, lightweight model can achieve superior performance over more complex architectures. This work underscores the potential for minimalist models to meet the needs of modern healthcare applications, particularly where resources are limited.
\end{abstract}

\section{Introduction}
Parkinson's disease (PD) is a common yet debilitating neurodegenerative disorder affecting 2\% of people over the age of 65 years\cite{scandalis2001resistance}. It is a progressive disorder that demands early and accurate diagnosis before a major loss of the dopaminergic neurons to improve patient outcomes\cite{poewe2017parkinson}. The diagnosis of PD is clinical and can be difficult without significant physical signs or symptoms. Electroencephalography (EEG) is a promising tool for non-invasive monitoring of subtle neurophysiological changes, offering potential for PD diagnosis. Recent studies have associated PD with various changes in EEG signals\cite{silberstein2005cortico,han2013investigation}. However, the complexity of EEG data presents significant challenges that require advanced machine learning models for detecting PD. Most existing deep-learning methods for PD classification using EEG data are complex and computationally expensive, which can hinder their practical application, especially in resource-constrained environments.

In this work, we propose LightCNN, a lightweight Convolutional Neural Network (CNN) architecture designed for efficient and effective classification of PD using EEG data. LightCNN stands out for its simplicity, utilizing just a single convolutional layer to achieve high performance. The motivation behind LightCNN is rooted in the principle that simplicity can be powerful. By focusing on essential features and reducing computational overhead, LightCNN not only offers high accuracy but also ensures interpretability and ease of deployment, especially in environments with limited resources, such as mobile and embedded systems.

To benchmark the performance of LightCNN, we compare it against several established deep learning architectures using EEG data from 46 participants, comprising 22 individuals with PD and 24 healthy controls. The results demonstrate that LightCNN not only rivals but exceeds the performance of more complex architectures, offering a powerful and computationally efficient alternative for PD classification. This makes LightCNN a promising candidate for real-time applications and resource-constrained environments, where both accuracy and efficiency are critical. Furthermore, we analyze the features captured by LightCNN and demonstrate its ability to identify neurophysiological patterns in EEG signals that are clinically relevant to PD. This capability, combined with its computational efficiency, positions LightCNN as a valuable tool for both research and clinical applications in neurodegenerative disease detection such as PD.

The rest of the paper is organized as follows. Section \ref{sec:priorworks} discusses prior deep learning approaches  for EEG-based classification of PD in the literature. Section \ref{sec:lightcnn} provides a detailed architecture and methodology of our proposed method, LightCNN. Section \ref{sec:exp} details our experiments  and the outcomes of our results are given in Section \ref{sec:results}. The ablation study is provided in Section \ref{sec:ablation}. Finally, Section \ref{sec:limitation}  is the discussion, and Section  \ref{sec:conclude} concludes the paper.

\section{Related Works}\label{sec:priorworks}

There are several deep learning models proposed in the literature for PD-related classification tasks using EEG ranging from detecting medication states of PD patients \cite{shah2020dynamical}, classification of PD during a specific task \cite{shi2019hybrid} or during resting state \cite{lee2019deep,lee2021,deepcnn} and discovering EEG biomarkers of PD \cite{vanegas2018machine}. In this study, we focused on PD classification using EEG data from the resting state. In this category, Oh et al. \cite{deepcnn} proposed a 13-layer CNN that achieved an accuracy of 88.3\%, a sensitivity of 84.7\%, and a specificity of 92\% using EEG data from 20 PD and 20 control subjects. Lee et al. proposed a hybrid model \cite{lee2019deep} consisting of a CNN and long-short-term memory (LSTM) layer achieving accuracy of 96.9\%, precision of 100\%, and recall of 93.4\% from 20 PD and 21 healthy subjects. Later, they proposed a modified version \cite{lee2021} with CNN and gated recurrent unit (GRU) layers achieving 99.2\% accuracy, 98.9\% precision, and 99.4\% recall. 

Additionally, several vision-based architectures have been proposed for EEG-based PD classification tasks. For example, Loh et al. \cite{loh2021gaborpdnet} proposed GaborNet which consists of 2D-CNN layers and works on Gabor-transformed spectrogram images of EEG data (15 PD and 16 Controls) achieving an accuracy of 99.46\% for 3-class classification (healthy vs PD with and without medication). Shaban et al.\cite{shaban2022resting} proposed a 20-layer 2D-CNN model applied on Wavelet-transformed images of EEG and achieved 99.6\% in the 3-class classification problem mentioned above. Khare et al. \cite{khare2021pdcnnet} proposed PDCNNNet which has 4 layers 2D-CNN applied on image-transformed EEG data achieving 99.97\% accuracy. 
 
Apart from these, several deep learning architectures are proposed for various EEG-based classification tasks such as EEGNet \cite{eegnet} and ConvNets \cite{convnet}.

\section{LightCNN: A Single  Convolutional Layer}\label{sec:lightcnn}
The proposed model is a lightweight CNN architecture designed for efficient EEG-based classification tasks. Its architecture is straightforward yet effective, featuring a single convolutional layer followed by a pooling and a fully connected layer (Figure \ref{Fig1}). The design emphasizes simplicity and computational efficiency, making it suitable for applications where a balance between performance and resource constraints is necessary. The architecture is composed of the following layers:
\begin{figure}
	\centering
	\includegraphics[width=13.5cm,height=6.5cm,clip,keepaspectratio]{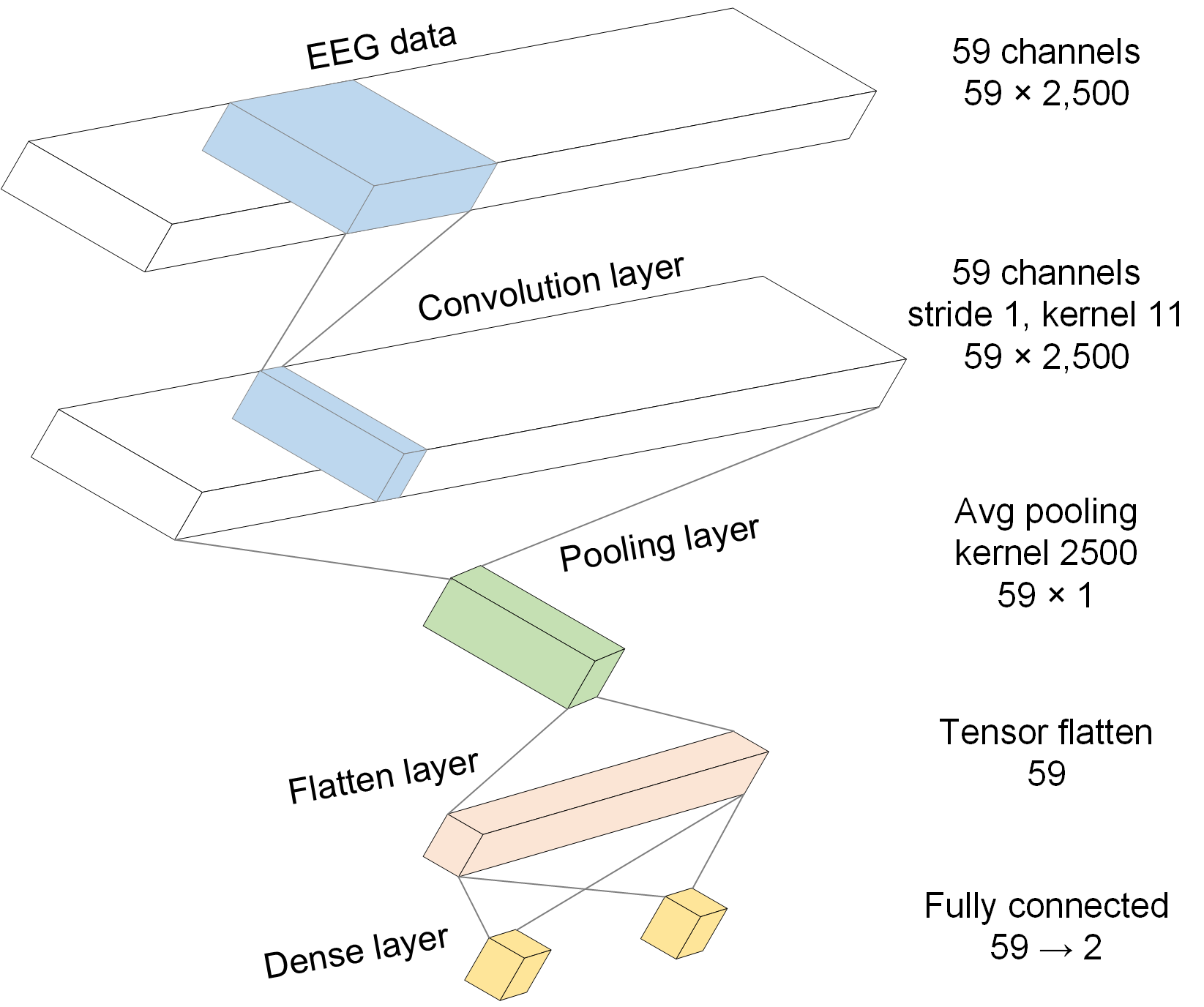}
	\caption{Overview of LightCNN model architecture.}
	\label{Fig1}
\end{figure} 

\begin{table}[htbp]
	\caption{Summary of the LightCNN model architecture.}
	\label{tab:cnn_model_summary}
	\centering
	\begin{tabular}{lllllll}
		\toprule
		Layer & Name  & Kernel size   & Stride    & Output shape & Parameters & Regularization \\ 
		\toprule
		0 & Input    & -  & -    & (59, 2500)           & -     & -              \\ 
		1 & Conv. 1D  & 11  &  1   & (59, 2500)           & 38,350  &     Dropout (0.1)        \\ 
		2 & AvgPool 1D  &  2,500  & 2,500  & (59, 1)              & -     & -              \\ 
		3 & FC & - & -     & (2,1)                  & 120    & -             \\
		\bottomrule
		\multicolumn{7}{l}{\footnotesize FC = Fully Connected layer; Conv. = Convolutional layer}\\ 
	\end{tabular}
\end{table}

\subsection{Convolutional Layer}
The input to the model is 1D signals from multiple EEG channels. The first layer is a 1D convolutional layer with the same number of output channels. The kernel size was 11. The layer applies padding to ensure the output has the same length as the input. This layer captures local dependencies in the signal by sliding the convolutional kernel across the time dimension of each channel. Following the convolution, the Rectified Linear Unit (ReLU) activation function is applied to introduce non-linearity into the model. A dropout layer with a dropout rate of 0.1 is then used to prevent overfitting by randomly setting a fraction of the input units to zero during training.

\subsection{Pooling Layer}
The output from the convolutional layer is passed through an average pooling layer with kernel size the same as signal length, which reduces the dimensionality of the data by taking the average over the entire length of the signal for each channel, resulting in a condensed representation of signal.

\subsection{Fully Connected Layer}
The pooled output is flattened and fed into a fully connected layer with two output nodes. Finally, a softmax function is applied to obtain classification output. The full model summary is given in Table \ref{tab:cnn_model_summary}.

\section{Experiments}\label{sec:exp}
\subsection{Dataset and Study Protocol}
We utilized a resting-state EEG dataset from 27 PD and 27 healthy participants which was collected during a study at the University of New Mexico (UNM; Albuquerque, New Mexico) \cite{anjum2020linear}. From these 54 participants, we utilized EEG data from 46 participants (22 PD and 24 healthy subjects) based on noise and artifacts through manual inspection of the data. PD patients were in OFF medication state.

\subsection{EEG Recording and Preprocessing}
The sampling rate ($F_s$) of EEG data was 500 Hz which was recorded by a 64-channel Brain Vision system. We utilized the first one minute of EEG data from each participant that corresponds to eyes closed resting state. EEG data from 59 channels out of 63 were utilized based on average channel data quality. Data from each channel were high-pass filtered at 1 Hz to remove noise and drift artifacts. No other pre-processing was implemented. Finally, the multi-channel data ($5n$ seconds) for each subject ($\mathbb{R}^{59\times 5n F_s}$) were segmented into 5-second epochs ($\mathbb{R}^{n\times 59\times 5F_s}$). These steps have been previously described in \cite{lipcot}.

\subsection{Experimental Setup}
We randomly shuffled data at the subject level and split the dataset into training (60\%), validation (20\%), and test (20\%) datasets. We utilized the training data for training the models. The validation data were used for evaluating the model's performance against overfitting and the best-performing model on the validation set was selected. To measure the classification performance, we utilized five metrics: precision, recall, accuracy, F1-score, and AUC. The classification performance was evaluated on the test dataset. 
 
\subsection{Performance Benchmarks}
To benchmark the performance of our proposed approach, we utilized five deep-learning architectures that have been shown to perform well in EEG-based classification tasks: 13-layer Deep CNN \cite{deepcnn}, ShallowConvNet \cite{convnet}, DeepConvNet \cite{convnet}, EEGNet \cite{eegnet} and Convolutional Recurrent Neural Network (CRNN) \cite{lee2021}. All methods were deep CNN architectures except for CRNN which utilized CNN and GRU layers. We chose these methods as they were shown to be very effective neural network architectures tailored for EEG-based PD classification in the literature. Model performances were evaluated on the test dataset while training and validation datasets were utilized for the training stage.

For performance comparisons, we chose end-to-end deep-learning approaches that take the raw EEG data as inputs and perform classification. We avoided approaches that require time-consuming feature extraction steps or image-transformation steps that generally rely on domain knowledge and human expertise. 
 
\subsection{Model Parameters}
During the training of LightCNN, the batch size was set to 2 with a learning rate of $1\times 10^{-4}$. Adam optimizer was utilized with a total of 80 epochs for the training.

\section{Results}\label{sec:results}
\subsection{LightCNN Outperforms State-of-the-art Methods}\label{results}
Our experimental results showed that our proposed model with a single convolutional layer outperformed the state-of-the-art architectures in all metrics (Table \ref{perf}). Among the five architectures compared in this study, CRNN provided the best overall performance. On the other hand, our LightCNN model outperformed CRNN by 2.3\% in recall, 4.6\% in precision, 0.1\% in AUC, 4\% in F1-score, and 3.3\% in accuracy. Note that CRNN employs a GRU layer which is computationally expensive.

\begin{table}[htbp]
	\caption{Performance comparison}
	\label{perf}
	\centering
	\begin{tabular}{llllllll}
		\toprule
		 Method  & Architecture & Layers & PRC  & Recall  & F1 & AUC & ACC \\
		\toprule
        DeepCNN\cite{deepcnn}  & CNN & 13 & 60.0      & 40.9   & 0.49     & 0.629 & 58.7   \\
         ShallowConvNet\cite{convnet} & CNN & 5 & 80.5      & 75.0   & 0.78     & 0.831 & 79.3   \\
         DeepConvNet\cite{convnet} & CNN & 14 & 87.8      & 81.8   & 0.85     & 0.917 & 85.9   \\
         EEGNet-8,2\cite{eegnet} & CNN & 11 & 88.6      & 88.6   & 0.89     & 0.967 & 89.1  \\
         CRNN\cite{lee2021} & CNN+GRU & 8 & 95.4      & 95.4   & 0.95     & 0.997 & 95.6   \\
		 \textbf{LightCNN (Ours)} & \textbf{CNN} & \textbf{3}  & \textbf{100}   & \textbf{97.7} & \textbf{0.99} & \textbf{0.998 }& \textbf{98.9 }\\
        
		\bottomrule
		\multicolumn{8}{l}{\footnotesize Best performance in bold. ACC = Accuracy, PRC = Precision}\\
	\end{tabular}
\end{table}

We also found that one of the deepest model with 13-layer CNN achieved the lowest performance. Apart from the exception of ShallowNet, there was an inverse relationship between the model depth (total layers in the architecture) and the performance metric, highlighting the fact that simpler models are better performing for EEG-based PD classification.

\subsection{Feature Interpretation: Unveiling the Success of LightCNN}\label{feature_inspect}
Our results demonstrated that LightCNN achieved superior performance compared to the state-of-the-art methods. However, we were also interested in understanding the underlying mechanism leading to this success. Unlike other methods, LightCNN adopts a very simple architecture which makes it possible to visualize and interpret different layers of LightCNN. After finalizing the model via training, we probed the model to investigate whether the EEG features detected by LightCNN had any clinical relevance to PD. In particular, we inspected the convolutional and pooling layer outputs and compared them with the pathological PD-related biomarkers present in the EEG dataset.

\subsubsection{PD-related Biomarkers in EEG Dataset}\label{psd_inspect}
\begin{figure}[htbp]
\centering
\includegraphics[width=14cm,height=10.5cm,clip,keepaspectratio]{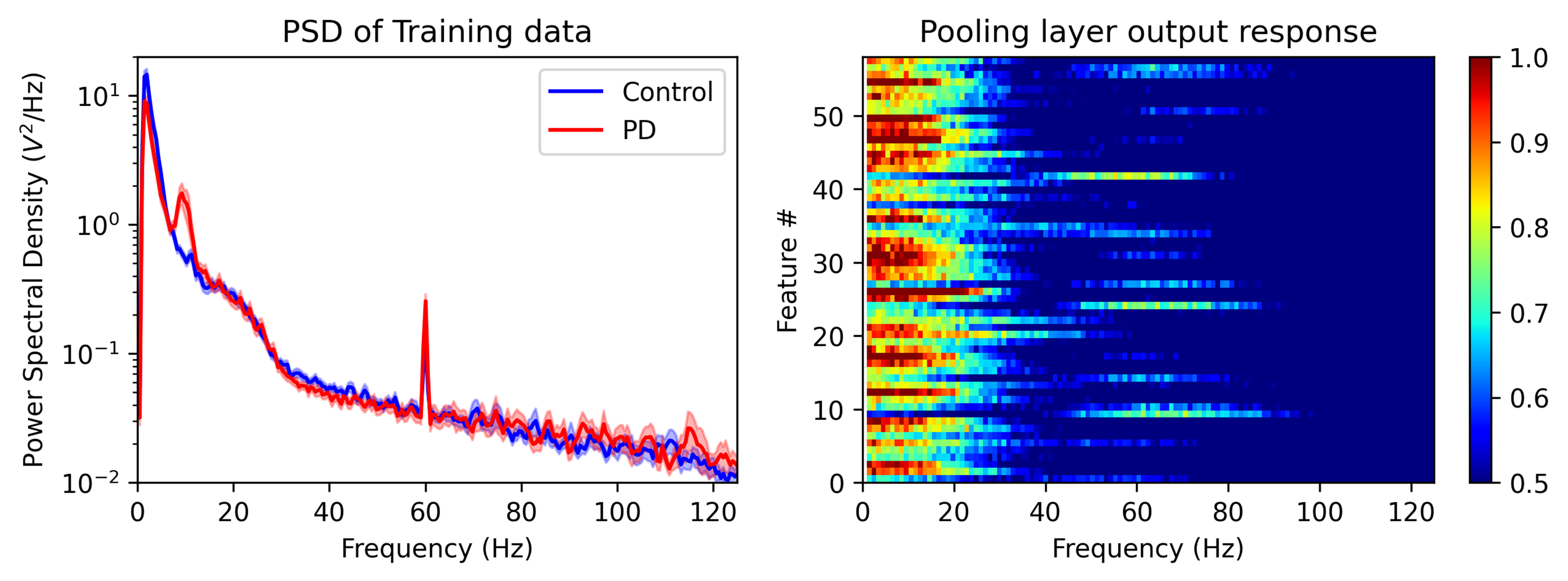}
\caption{Feature interpretation of LightCNN: Left panel shows power spectrum (PSD) comparison between PD and healthy controls (mean$\pm$ SEM) highlighting the PD-related changes in the frequency domain. Data from all channels were averaged before PSD calculation and 5s epochs were utilized. Right panel shows the frequency response of the pooling output layer of a trained LightCNN model where x-axis is frequency (Hz), y-axis represents pooling layer output or features and the colors show average activation value. }
\label{AF0}
 \end{figure}

First, we investigated the pathological biomarkers of PD in our EEG dataset to establish a reference point for our interpretation. One of the most widely used methods to analyze EEG signals is power spectral analysis (PSD). It is well-known that the key neurophysiological components of EEG data are in low-frequency range focused in several pre-defined frequency bands such as delta (0.1-4 Hz), theta (4-8 Hz), alpha (8-13 Hz), beta(13-30 Hz) and gamma (30> Hz) that are linked to different brain functionalities \cite{kumar2012analysis,groppe2013dominant}. Out of these, elevated beta rhythms among PD participants are well-established which correlates with the PD-related movement symptoms \cite{little2014}. Consequently, group-wise PSD of PD and healthy control participants in our training dataset showed an elevated beta peak in PD participants as well as elevated delta rhythm in healthy controls (Figure \ref{AF0}). Interestingly, there was a sharp artifact peak at 60 Hz that modulated with PD. 

\subsubsection{Inspecting the Pooling Layer Output}\label{pool_inspect}  
As the neurophysiological biomarkers of PD in EEG signals are mostly defined in the frequency domain, we investigated the sensitivity to various frequency components of the pooling layer output which is also the input nodes of our fully connected layer. To achieve this, we generated single-tone sinusoidal input signals for our LightCNN models with a single frequency component and observed the activation of the pooling output layer to measure the sensitivity for the given frequency (Appendix \ref{ap1}). We measured these sensitivities for all frequencies and compared the results. Figure \ref{AF0} shows the frequency sensitivity of the pooling output layer which highlights that most of the pooling outputs were sensitive to the low-frequency range (<30 Hz) while many were also sensitive to the gamma range near 60 Hz (40-80 Hz). These indicate that the pooling layer was particularly sensitive to the key brain rhythms (delta, theta, alpha, beta) as well as the PD-related artifact modulation near 60 Hz.

\subsubsection{Inspecting the Convolutional Layer Output}\label{conv_inspect}
\begin{figure}[htbp]
\centering
\includegraphics[width=14cm,height=10.5cm,clip,keepaspectratio]{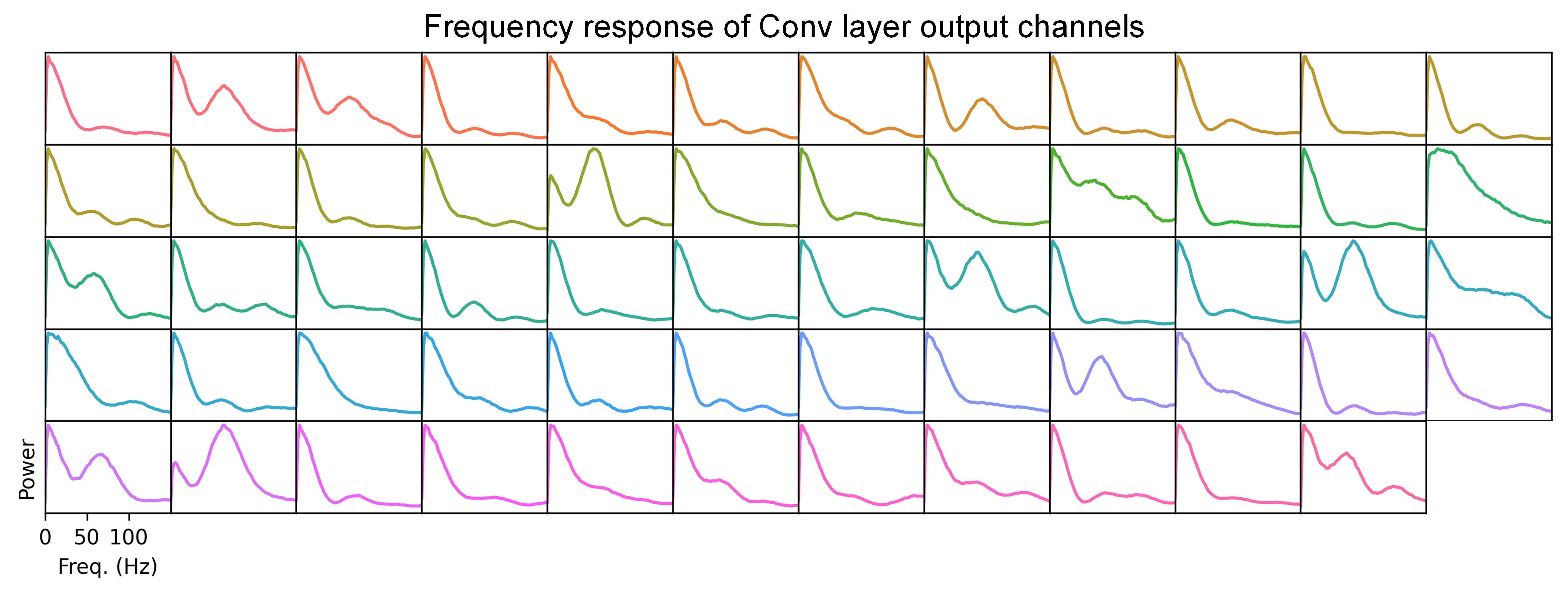}
\caption{Frequency response of the  convolutional layer output channels: Each pallet shows the filtering profile of a single  convolutional output channel where x-axis is frequency in Hz and y-axis is power in log scale. All 59 output channels are shown from a trained LightCNN model. }
\label{AF1}
 \end{figure}

Finally, we investigated the frequency components captured by the  convolutional layer of LightCNN. Note that convolution operation is analogous to a filtering process. Therefore, we were interested in the frequency response of the  convolutional layer's filtering processes to compare with the pathological frequency markers of PD. To achieve this, we generated white noise that has a flat PSD profile (same power in every frequency range) and utilized this as input to the already-trained LightCNN model (Appendix \ref{ap1}). Then, we observed the PSD of the  convolutional layer's output channels which represent the frequency response of the filtering processes. Figure \ref{AF1} shows the frequency response of the  convolutional layer's output channel from a trained LightCNN model which illustrates that the majority of the output channels were filtering the low-frequency range (<30 Hz) contents. Additionally, a relatively broadband filtering at the gamma range was also present in many of the output channels. These results show that similar to the pooling layer, the  convolutional layer also captured the neurophysiological components of EEG  relevant to PD while simultaneously capturing the 60 Hz artifact modulation.

Combining insights from above, we found that the  convolutional layer functions as a filter bank utilizing 59 filters to select frequency components of EEG data in both the low-frequency range (< 30 Hz) and the gamma range around 60 Hz, both of which are clinically relevant to the pathological EEG neurophysiology of PD. The subsequent pooling layer then averages these filtered signals, effectively quantifying the signal strength within the selected frequency bands. Finally, the fully connected layer assigns weights to these signals to produce the final classification outcome.

\section{Ablation Study}\label{sec:ablation}
\subsection{Effect of Kernel Size}
\begin{figure}[htbp]
\centering
\includegraphics[width=13.5cm,height=10.5cm,clip,keepaspectratio]{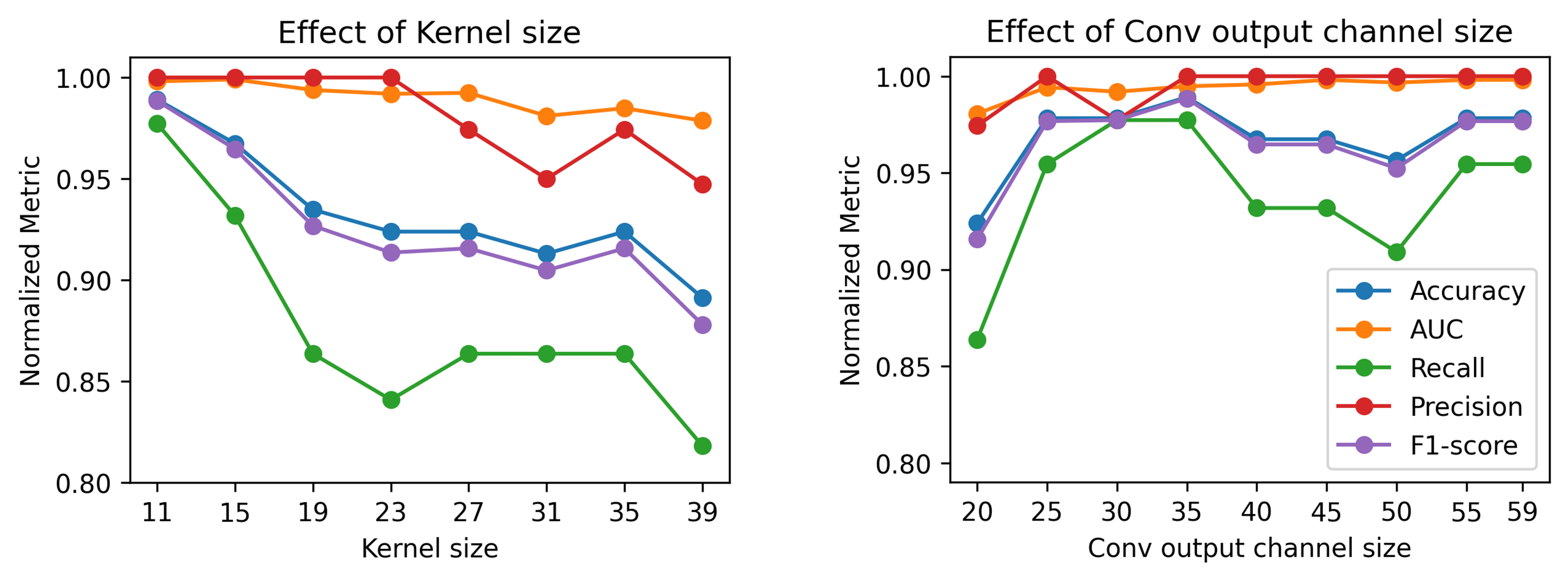}
\caption{Ablation study: Evaluating LightCNN's performance while varying the kernel size  (left) and the output channel size (right) of the  convolutional layer. All performance metrics were normalized to the range 0 to 1. Performance measured on the test dataset.}
\label{AF2}
 \end{figure}

First, we investigated the influence of the kernel size of the  convolutional layer on the performance of our LightCNN architecture. For this, we varied the kernel size from 11 to 39 and evaluated the model's performance on the test dataset. All other parameters were fixed. Figure \ref{AF2} shows the model's performance for varying kernel size. In our experimental results, we observed that the smallest kernel size (11) provided the best performance. Additionally, there was a noticeable decline in performance as the kernel size increased, indicating that larger kernels may lead to reduced efficacy in feature extraction. This trend suggests that smaller kernels are more suitable for optimizing the performance of LightCNN in EEG-based PD classification. Notably, while most performance metrics deteriorated with larger kernels, the AUC exhibited minimal change, whereas recall metrics showed the greatest variability. This implies that models with larger kernels may develop class-specific biases, resulting in less balanced classification outcomes.

\subsection{Effect of  Convolutional Layer Output Channel Size}
Next, we explored the effect of varying the output channel size of the convolutional layer. We systematically adjusted the output channel size from 20 to 59 and assessed the model's performance. Our results showed that, unlike kernel size, output channel size showed no consistent impact on the model's performance (Figure \ref{AF2}). In particular, AUC and precision showed minimal changes across different output channel sizes, suggesting that this parameter has less influence on the overall effectiveness of LightCNN.

\section{Discussion}\label{sec:limitation}
In this study, we introduced LightCNN, a streamlined yet highly effective CNN architecture tailored for PD classification using EEG data. Despite its simplicity, our model with a single  convolutional layer demonstrated remarkable performance, surpassing state-of-the-art (SOTA) architectures across all evaluation metrics. This achievement underscores the potential of minimalist architectures to deliver superior results without the computational complexity often associated with more sophisticated models.

Among the SOTA architectures compared, CRNN emerged as the closest competitor, providing strong overall performance. However, our LightCNN model outperformed CRNN by significant margins: a 2.3\% improvement in recall, a 4.6\% increase in precision, a 4\% boost in F1-score, and a 3.3\% higher accuracy. These results highlight LightCNN's effectiveness in achieving a balanced and robust performance in all metrics. CRNN's use of a GRU layer, while effective, introduces substantial computational demands and challenges in scalability, particularly for large-scale or real-time applications. In contrast, LightCNN's architecture avoids these complexities, offering a more efficient and scalable alternative without compromising on performance. The simplicity of our approach not only facilitates easier implementation but also makes it more adaptable to scenarios where computational resources are constrained. 

Our investigation of LightCNN's features showed that it can capture the neurophysiological changes of EEG that have clinical relevance in PD. Specifically, the  convolutional layer operates as a filter bank, isolating critical frequency components relevant to PD, while the pooling layer evaluates the signal strength within these frequencies. The fully connected layer then weights these frequency-specific signal strengths to perform the classification. The high classification performance of this simple architecture shows that indeed the frequency-specific contents captured by a single  convolutional layer are powerful enough to analyze EEG signals for PD classification. 

Our findings have important implications. First, they demonstrate that a well-designed CNN architecture can effectively capture the necessary features for PD classification from EEG data, eliminating the need for more complex networks for EEG analysis. Indeed, our results showed an inverse relationship between the model's depth of layers and performance.  Furthermore, our ablation study showed that the simpler version of LightCNN with a lower kernel size is better for PD classification.  These indicate that while EEG data are complex in nature, this complexity does not warrant deep neural architectures. On the contrary, larger models tend to overfit and memorize EEG data. They also need larger training datasets which is not suitable for EEG-based analysis due to the scarcity of such datasets.  Second, the performance gains achieved by LightCNN suggest that lightweight models can be both efficient and powerful, making them suitable for deployment in resource-limited environments, such as mobile or embedded systems. Third, simpler models are more interpretative and the features captured by such models have clinical relevance which is a key requirement in medical applications.

Future research could explore the generalizability of LightCNN to other neurodegenerative disorders or broader EEG-based classification tasks. Additionally, integrating techniques like model quantization or pruning could further enhance LightCNN's efficiency, making it an even more attractive option for real-time and edge computing applications.

\section{Conclusion}\label{sec:conclude}
In conclusion, our study represents a compelling argument that LightCNN, a simple CNN architecture with a single  convolutional layer offers both accuracy and efficiency in EEG-based PD classification. Our results show that it can outperform more complex architectures and effectively capture clinically relevant neurophysiological changes in EEG, while maintaining computational efficiency. These findings position LightCNN as a valuable tool for both research and clinical applications in the field of neurodegenerative disease detection.
 
\subsection*{Data and Code Availability}
The original EEG dataset can be found at \href{http://predict.cs.unm.edu/downloads}{http://predict.cs.unm.edu/downloads.php}. The pre-processed EEG dataset in .csv formats can be found via this Dropbox \href{https://www.dropbox.com/scl/fi/xinqn33vof0bnb9rlvmdh/raw.zip?rlkey=jb4dyumh7v82vbj36wsb53x13&dl=0}{link} and the Python codes are in \href{https://github.com/MDFahimAnjum/LightCNNforPD}{https://github.com/MDFahimAnjum/LightCNNforPD}.


\bibliographystyle{plain} 
\setlength{\itemindent}{0pt} 

\appendix
\section{Appendix}
\subsection{Signal Generation for Feature Interpretation}\label{ap1}
\begin{figure}[htbp]
 	\centering
 	\includegraphics[width=14cm,height=10.5cm,clip,keepaspectratio]{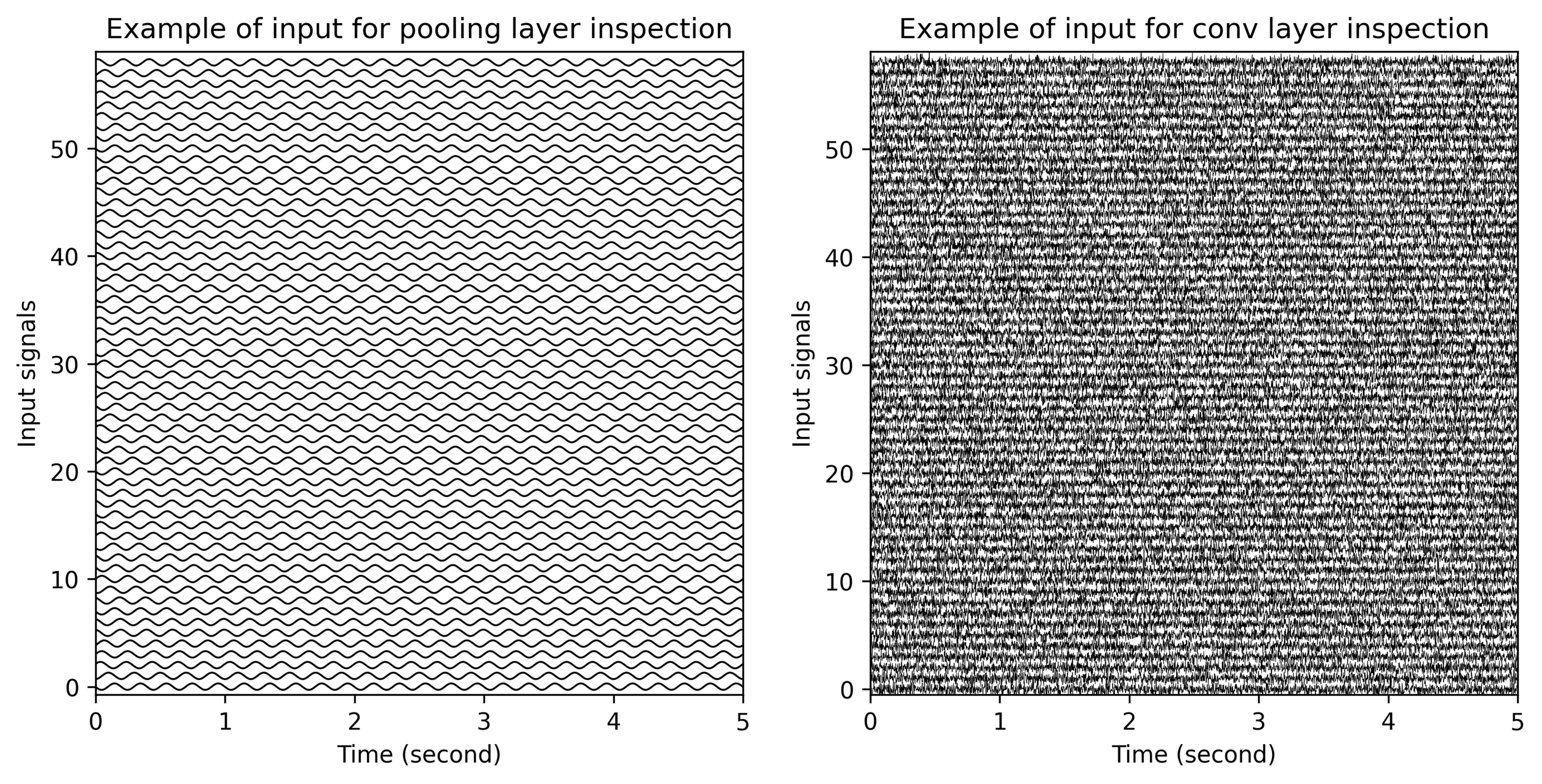}
 	\caption{Example synthetic data for measuring pooling layer sensitivity (left) for 5 Hz frequency and for evaluating filtering responses of the  convolutional layer output channels (right). x-axis is time in seconds and y-axis shows 59 channels.}
 	\label{Aa1}
 \end{figure}

For probing the frequency response of the pooling output layer, we generated synthetic single-tone sinusoidal signals which were fed to the trained LightCNN model instead of EEG data. Note that the EEG data for each 5-second epoch consists of 59 time series (one for each channel) each with a length of 2,500 samples (5-second data with 500 Hz sampling rate). Similarly to obtain synthetic data for probing the model, we generated 59 time series data each having a single frequency component ($f$) with a constant amplitude of 1 and a random phase. The addition of a random phase ensured that the channel data were not identical, as there were random shifts in the data while the sinusoids had the same frequency and amplitude. The sinusoidal data for channel $n$ for a given time range $t\in \{0,\nicefrac{1}{F_s},\nicefrac{2}{F_s},\dots,5\}$ is,
\begin{equation}
	x_n=A \mbox{sin}(2\pi f t +\phi_n)
\end{equation}
where, $A=1$ and $\phi_n\sim\mathcal{U}[0,2\pi]$. Note that $\nicefrac{F_s}{2}$ Hz is the Nyquist frequency. Figure \ref{Aa1} shows an example 5-second epoch of synthetic data. We provided these data as input to the model and obtained the output data from the pooling layer. This process was repeated for multiple frequencies ( $f=\{0,1,,\dots,\nicefrac{F_s}{2}\}$). Finally, these steps were repeated several times ($n=100$) and the pooling layer outputs were averaged to obtain the frequency response.

 To obtain the filtering profile of the  convolutional output channels, we generated White noise with a flat power spectrum as inputs for all channels. This resulted in 59 time series data of white noise with a length of 2,500 samples (Figure \ref{Aa1}) which were given as input to the model. The whole process was repeated multiple times ($n=300$) and the output data from the  convolutional channels were collected. Finally, we calculated PSD from the output data to obtain the filtering profile of the output channels.
  
\end{document}